# Research and Design on Intelligent Recognition of Unordered Targets for Robots Based on Reinforcement Learning

Yiting Mao[1*], Dajun Tao[2]，Shengyuan Zhang[3]，Tian Qi[4]，Keqin Li[5]
[1]Hefei University of Technology，China
[2]Carnegie Mellon University, School of Engineering, USA
[3]Cornell University, College of Computing and Information Science, USA
[4]University of San Francisco, College of Arts and Sciences,USA
[5] AMA University Quezon, Department of Computer Science, Philippines
[1*]2070656994@qq.com, [2]dajunt@alumni.cmu.edu, [3]sophiazhang217@gmail.com, [4]tqi51212@gmail.com,
[5]keqin157@gmail.com

**Abstract:** In the field of robot target recognition research driven by artificial intelligence (AI), factors such as the disordered distribution of targets, the complexity of the environment, the massive scale of data, and noise interference have significantly restricted the improvement of target recognition accuracy. Against the backdrop of the continuous iteration and upgrading of current AI technologies, to meet the demand for accurate recognition of disordered targets by intelligent robots in complex and changeable scenarios, this study innovatively proposes an AI - based intelligent robot disordered target recognition method using reinforcement learning. This method processes the collected target images with the bilateral filtering algorithm, decomposing them into low - illumination images and reflection images. Subsequently, it adopts differentiated AI strategies, compressing the illumination images and enhancing the reflection images respectively, and then fuses the two parts of images to generate a new image. On this basis, this study deeply integrates deep learning, a core AI technology, with the reinforcement learning algorithm. The enhanced target images are input into a deep reinforcement learning model for training, ultimately enabling the AI - based intelligent robot to efficiently recognize disordered targets. Experimental results show that the proposed method can not only significantly improve the quality of target images but also enable the AI - based intelligent robot to complete the recognition task of disordered targets with higher efficiency and accuracy, demonstrating extremely high application value and broad development prospects in the field of AI robots.

# 1 Introduction

In the era of the booming development of artificial intelligence (AI), intelligent robots, as a typical representative of cutting-edge technology, have become the crystallization of the intersection and integration of multiple disciplines. It is a highly integrated intelligent device that can deeply integrate various advanced functions such as environmental perception, dynamic decision-making, and autonomous learning into one in a completely unknown and complex and changeable environment, constructing a complex and precise intelligent system. The research and development process of intelligent robots deeply depends on the collaborative innovation of a series of cutting-edge technologies such as artificial intelligence, information fusion, machine learning, and pattern recognition. Among them, artificial intelligence technology endows intelligent robots with human-like perception, understanding, and decision-making abilities; information fusion technology is responsible for integrating massive data from different sensors, eliminating data redundancy and contradictions, and providing a reliable basis for the accurate decision-making of robots. With the support of these advanced technologies, intelligent robots have shown extremely broad application prospects and development potential in many fields such as military, aerospace, industrial manufacturing, and medical services [1].

Among the numerous functions possessed by intelligent robots, target recognition undoubtedly occupies a pivotal core position. The ability of target recognition is the basis for intelligent robots to achieve autonomous actions and task execution. Intelligent robots use various high-precision sensors carried by themselves, such as cameras, lidars, and infrared sensors, to conduct all-round and multi-level perception processing of the targets in the surrounding environment. These sensors transmit the collected raw data to the central processing unit of the robot, and through a series of complex AI algorithms, the classification, positioning, and recognition of the targets are realized, thus providing key information for subsequent decision-making and actions[2].

In recent years, many domestic experts and scholars have carried out a large number of in-depth and fruitful research works around the key field of intelligent robot target recognition. Feng Zhiguang et al[3]. focused on the application of robots in the agricultural picking field and proposed a target recognition method for picking robots based on Softmax.This method makes full use of the Softmax classification model in artificial intelligence. As a classic multi-classification algorithm, this model can perform probabilistic processing on the input feature vectors, thereby achieving accurate classification of different picking targets. By constructing and optimizing the Softmax classification model, the efficient recognition of fruits, crops, and other targets by picking robots has been successfully achieved, providing strong support for the intelligence and automation of agricultural production.

Zhao Pengyu et al. [4] focused on the Delta robot in industrial production and proposed a target recognition algorithm based on machine vision. This method relies on the machine vision technology in artificial intelligence. First, it performs systematic calibration on the Delta robot, uses advanced mathematical models and algorithms to accurately determine the position transformation relationship among the camera, the robot, and the conveyor belt, and establishes a precise spatial coordinate mapping system. On this basis, the ORB (Oriented FAST and Rotated BRIEF) feature extraction algorithm is organically combined with the Onecut algorithm. As an efficient feature extraction algorithm, the ORB algorithm can quickly and accurately extract the key feature points in the image; the Onecut algorithm is used to further screen and process the extracted features, thereby achieving the precise recognition of robot targets and effectively improving the work efficiency and accuracy of the Delta robot on the industrial production line.

As an advanced instance segmentation algorithm, SOLOv2 has high accuracy and efficiency in the field of target recognition. Wu You et al. made targeted improvements to the SOLOv2 algorithm to make it better adapt to the target detection and recognition requirements in complex scenarios. The improved algorithm makes full use of the deep learning technology in artificial intelligence. Through training with a large number of sample data, intelligent robots can accurately identify the status of various equipment, fault points, and potential safety hazards in complex industrial environments, power facility inspections, and other scenarios, playing an important role in ensuring the stable operation of the system.

However, although the above existing methods have achieved certain results in their respective application fields, some problems to be solved are still exposed in practical applications. For example, in the face of complex situations such as the disordered distribution of targets, drastic changes in the environment, noise interference in data, and real-time requirements, the recognition accuracy and stability of these methods often fail to meet the actual needs. In order to overcome these problems and further improve the performance of intelligent robot target recognition, so that it can better adapt to complex and changeable practical application scenarios, this research innovatively proposes an intelligent robot disordered target recognition method based on reinforcement learning. As an important branch of the artificial intelligence field, reinforcement learning can enable intelligent robots to autonomously explore the optimal target recognition strategy through continuous trial and error and learning in the interaction process with the environment. Experimental results show that this method can significantly improve the accuracy of the recognition results of intelligent robots for disordered targets, still maintain high recognition accuracy and stability in complex environments, has high application value and practical significance, and provides new ideas and methods for the development of intelligent robot target recognition technology.

# 2 Methods

## 2.1 Target Image Enhancement for Intelligent Robots

The target images are collected by the intelligent robot. However, due to various factors such as illumination during the collection process, the obtained target images do not show significant effects, and thus enhancement processing of the target images is required [5.

The target image is regarded as the product of the illumination image H（x,y）and the reflection image Z(x,y), and the corresponding expression is as follows:

$$L(x,y) = H(x,y) \times Z(x,y) \quad (1)$$

where L(x,y) represents the target image.

In the actual processing, logarithmic transformation is first used to convert the product relationship between different images into a summation relationship, and at the same time, the images are decomposed. In addition, during the process of illumination estimation, the distance between the pixels of the target image itself and its adjacent pixels is comprehensively considered to avoid the influence of excessively high or low pixel values on the image, and also to prevent the occurrence of halation in the target image.

The original target images collected by the intelligent robot are input into the logarithmic domain. According to the illumination estimation results, they are divided into two parts, and different methods are used to process these two parts respectively. The two processed parts are combined to form an image within the hologram.

The illumination is estimated through bilateral filtering. Due to various advantages of the bilateral filtering algorithm, it is widely applied in the field of image processing. During the bilateral processing of the target image, the correlation between the output value of the target image and the spatial position of the surrounding pixels needs to be considered, which is expressed in the form of Equation (2):

$$Z_{(s)} = \frac{1}{c(x)} \sum_{k \in \psi} L(x,y) s(e-r) u(l_p - l_s) l_p \quad (2)$$

In this equation, Z(s) represents the output value of the bilateral filtering result. ψ represents the set of pixels corresponding to the target image. and denote two Gaussian functions with different values. and represent the spatial domain and the luminance domain respectively. $l_p$ and $l_s$ are the weight contributions of the spatial domain and the luminance domain respectively. c(x) represents the normalization factor, which needs to satisfy:

$$c(x) = \sum_{k \in \psi} s(e-r) u(l_p - l_s) \quad (3)$$

In the actual implementation process, if bilateral filtering is only performed based on the original theory, the computation time is relatively long. To effectively address the above-mentioned problem, a method of grayscale value layering is introduced to improve the computational efficiency of the bilateral filtering algorithm. The detailed calculation steps are as follows:

1.After the acquisition and processing of the target image, it is necessary to initialize the vector grid $\beta(z_x, z_y, t)$ corresponding to the target image, and at the same time, it should satisfy:

$$\beta(z_x, z_y, t) = \begin{cases} (E(z_x, z_y), 1), if & t = E(z_x, z_y) \\ (0,0,0), otherwise \end{cases} \quad (4)$$

In the formula, $E(z_x, z_y)$ represents a two-dimensional grayscale image, and represents the running time.

2.Perform Gaussian filtering on each layer of the vector grid separately. The corresponding calculation formula is as follows:

$$U_{\beta(z_x,z_y,t)} = Z_{(s)}H_{(\sigma,\sigma)} * \beta(z_x,z_y,t) \quad (5)$$

In the formula, H($\sigma$, $\sigma$) represents a three-dimensional Gaussian function.

After the above operations, if the final result at the position ($z_x$,$z_y$,t) is set as ($h_i$,$h_j$), then the calculation result U($\beta$) of the bilateral filtering is shown as in Equation (6):

$$U_{(\beta)} = \frac{(h_i,h_j)^2}{\beta(z_x,z_y,t)} \quad (6)$$

After the illumination estimation of the target image is achieved, the illumination image can be obtained. A part of the pixels are respectively intercepted at both ends of the illumination image by the histogram interception method, and the value range of the remaining pixels is set. At the same time, the illumination image is corrected by the improved Gamma correction method, and relevant operations such as linear stretching are carried out on the corrected illumination image. The improved Gamma correction is expressed in the form of Equation (7):

$$s(i) = U_{(\beta)} y^{d*a+a} \quad (7)$$

In the formula, s(i) represents the output value, y represents the pixel value of the target image, d represents the control parameter, and a represents the correction parameter.

The illumination compression processing of the target image is carried out through the improved Gamma correction formula, and the corresponding expression is as follows:

$$W_{(x,y,z)} = (s(i) * y^{d*a+a}) - (h_i, h_j) \quad (8)$$

In the formula, W(x,y,z) represents the target image after illumination compression processing.

After obtaining the illumination image through the above operations, it is necessary to subtract the illumination image from the original target image in the logarithmic domain, thereby obtaining the reflection image. The Sigmoid function is used to enhance the reflection image, and the detailed operation steps are as follows:

1. Calculate the luminance value l(u) of the target image:

$$l(u) = W(x,y,z)(0.2989 \times r + 0.587 \times g + 0.114 \times b) \quad (9)$$

where r,g and b represent different color channels respectively.

2. Take the logarithm.
3. Process the target image by combining the bilateral filtering algorithm with the grayscale-value layering acceleration technique to obtain the illumination image H(x,y).
4. Subtract the luminance image l(x,y) from the illumination image H(x,y) to obtain the reflection image as shown in Equation (10):

$$Z(x,y) = l(x,y) - H(x,y) \quad (10)$$

5. Use the histogram truncation method to truncate a certain number of pixels at the left and right ends of the illumination image respectively, and perform enhancement processing I(r) through the Sigmoid function. The corresponding expression is as follows:

$$I(r) = \frac{2Z(x,y)}{1+\beta_{(a,b)}} - 1 \quad (11)$$

In the formula, represents the mapping function.
After the enhancement process, a stretching operation also needs to be carried out on it.

6. By adding the illumination image and the reflection image, a new image New(x,y) can be obtained, achieving the enhancement of the target image of the intelligent robot. The corresponding calculation formula is as follows:

$$New(x,y) = I(r)[H(x,y) + Z(x,y)] \quad (12)$$

## 2.2 Unordered Target Recognition of Intelligent Robots Using Reinforcement Learning

Since deep learning and reinforcement analysis possess powerful comprehension capabilities and precise decision-making capabilities respectively, combining the two effectively to form deep reinforcement learning can ensure better learning between each end-to-end. Unordered target recognition of intelligent robots is carried out through deep reinforcement learning. The schematic diagram of the principle of deep reinforcement learning is shown in Figure 1.

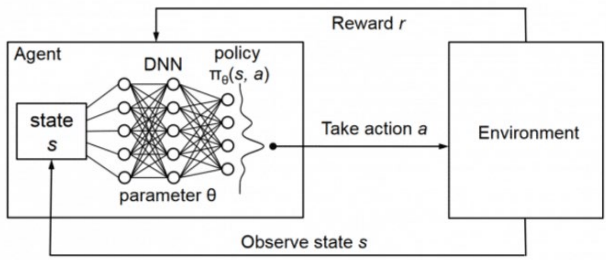

**Figure 1.Schematic diagram of the principle of deep reinforcement learning**

In the field of intelligent robot target recognition, the process of achieving unordered target recognition based on the Faster Region - based Convolutional Neural Network (Faster R - CNN) mainly undergoes two crucial stages. First, in the feature extraction stage, a specific convolutional neural network structure is utilized to conduct in - depth feature mining and extraction on the input image data, thereby obtaining key feature information that can characterize the target objects in the image. Subsequently, based on the extracted features, a Region Proposal Network (RPN) is constructed. Through a series of complex calculations and screenings, this network generates a series of candidate regions that may contain target objects on the feature map, and then determines the Regions of Interest (RoI) from them.

On this basis, to further enhance the accuracy and efficiency of unordered target recognition by intelligent robots, this study, based on the Faster R - CNN architecture, innovatively proposes a more efficient unordered target recognition method for intelligent robots based on deep reinforcement learning. Specifically, the advanced machine - learning technique of reinforcement learning is employed to conduct targeted optimization of the Region Proposal Network (RPN). By continuously adjusting the network parameters, the RPN can generate high - quality candidate regions more accurately. At the same time, the reinforcement function in reinforcement learning is carefully improved to ensure that the proposal - forming effect in each region is significantly enhanced, thereby improving the accuracy of target recognition. In addition, to effectively reduce the computational complexity of the algorithm, the lightweight convolutional neural network MobileNet is introduced for feature extraction. The unique network structure design of MobileNet can, while ensuring the feature - extraction effect, significantly reduce the amount of computation and improve the operational efficiency of the algorithm.

During the actual calculation process, the agent in reinforcement learning needs to pre - set strict criteria for being selected as an RoI, as follows:

1.The agent needs to facilitate the generation of object instances with high confidence and perform overlapping operations on these instances to more comprehensively capture the feature information of the target objects. Here, the overlapping operation refers to setting a reasonable overlap among the multiple generated candidate regions to ensure that different parts of the target object can be covered, thereby improving the ability to fully recognize the target object.

2.The agent needs to ensure that the number of generated RoIs is minimized while effectively avoiding false alarms. This requires the agent, when generating candidate regions, to not only consider the quantity of regions but also pay more attention to the quality of regions, that is, to minimize the number of false candidate regions that do not contain real target objects, thereby improving the accuracy of recognition.

Considering that target images usually exhibit characteristics such as varying sizes and relatively large target instances, for each instance object, in the initial stage, the fixation - action reward mechanism will assign a small negative reward to each fixation action. This setting aims to guide the agent to try different fixation strategies during the initial exploration stage and avoid prematurely falling into a local optimal solution. As the agent explores, it will obtain an increasing Intersection over Union (IoU) value composed of random instances in the current target image. When the IoU value increases, it means that the overlap degree between the candidate region selected by the agent and the real target object is improved, indicating that the agents decision - making is more accurate. At this time, the agent will receive a positive reward. The fixation reward at time t after the adjustment process can be calculated through Equation (13), thereby continuously optimizing the agents decision - making strategy and improving the performance of unordered target recognition by intelligent robots.

$$f_t^k = \alpha New(x,y) + \frac{1}{\rho}\left(\sum IoU_t^i - IoU_t^j\right) \quad (13)$$

In the formula, $f_k^t$ represents the fixation reward at time t; α represents the probability vector; ρ represents the number of bounding boxes of a specific class; $IoU_i^t$ and $IoU_j^t$ represent the Intersection over Union of the i-th instance object and the j-th instance object respectively.

After obtaining the fixation reward, it is also necessary to calculate the IoU between different instance objects and the ground truth, as well as the covered area. Through analysis, it can be seen that the larger the covered area, the closer the value of the reward is to 0; otherwise, the reward will become negative. At the same time, the agent will receive the completion action reward $f_d^t$ that can accurately describe the change in the quality of the search trajectory. The corresponding calculation formula is as follows:

$$f_t^d = \frac{1}{\rho}\left(\sum IoU_t^i - \rho\right) \quad (14)$$

Deep reinforcement learning is designed based on the VGG-16 network structure. Although the overall structure of VGG16 is not extremely complex, compared with similar models, VGG16 has a relatively high accuracy. However, during the calculation process, it requires a large amount of data and takes a relatively long time. To effectively address the above-mentioned issues, it is necessary to construct the lightweight convolutional neural network Mobile-Net V1 and determine the number of network layers. Among them, the standard convolutional kernel and the depth-separable convolutional kernel are applied to the first layer and other layers of the deep reinforcement learning network respectively. In Mobile-Net V1, a certain amount of data and parameters need to be integrated and placed on the convolutional layer uniformly, and some data and parameters are placed on the fully-connected layer.

Through the above analysis, for the unordered target recognition operation of intelligent robots using deep reinforcement learning, the Mobile-Net V1 network is used to replace the VGG16 network. At the same time, the enhanced target images are input into the deep reinforcement learning for relevant testing and training operations. In this process, not only does it need to effectively improve the accuracy of the recognition results, but also adjust the parameters of the lightweight convolutional neural network to effectively reduce the network training time and the number of parameters, so as to obtain a training model with better performance.

After completing the above operations, all the enhanced target images are input into the deep reinforcement learning network for training, thereby establishing the target image training matrix, as shown in Equation (15):

$$P = \begin{bmatrix} p_{11}, p_{12}, p_{13}, \cdots, p_{1n} \\ p_{21}, p_{22}, p_{23}, \cdots, p_{2n} \\ \vdots \quad \vdots \quad \vdots \\ p_{m1}, p_{m2}, p_{m3}, \cdots, p_{mn} \end{bmatrix} \quad (15)$$

Effectively combine the deep reinforcement learning model with the actual recognition task, conduct classification and recognition through a classifier, and ultimately achieve unordered target recognition for intelligent robots. The corresponding calculation formula is as follows:

$$S = \frac{1}{P}\left[f_t^k + f_t^d\right]\frac{1}{\rho}\left(\sum IoU_t^i - IoU_t^j\right) \quad (16)$$

## 3. Simulation Research

### 3.1 Enhancement Effect of Target Images

In order to verify the effectiveness of the proposed unordered target recognition method for intelligent robots using reinforcement learning (hereinafter referred to as "the proposed method"), some images captured by intelligent robots are selected as test images for simulation. The target recognition method based on Softmax in Ref. [3] (hereinafter referred to as "the method in Ref. [3]") is adopted as a comparative method for experimental testing. Through the analysis of experimental data, it can be seen that after the test images are enhanced by the proposed method, the contrast, brightness, etc. of the images are significantly improved. Meanwhile, the edge details of the images become clearer, and the visual effect is better.

To further verify the quality of the enhancement effect, traditional image enhancement algorithms are selected as comparative methods, and the following indicators are selected for test analysis:

Information entropy: The larger its value, the richer the content contained in the image.

Mean value: The larger its value, the better the overall brightness of the image.

Average gradient value: The larger its value, more levels can be reflected, and the clarity of the image is also better.

## 3.2 Unordered Target Recognition Time of Intelligent Robots

The experimental results of the unordered target recognition time of intelligent robots using different methods are shown in Table 1.

**Table 1 Comparison of experimental results of target disorder recognition time of intelligent robots with different methods**

| Test target number | Unordered target recognition time of the method in Ref. [3] / (ms) | Unordered target recognition time of the method in this paper / (ms) |
|---|---|---|
| 1 | 6.6 | 5.2 |
| 2 | 7.1 | 6.3 |
| 3 | 10.5 | 8.9 |
| 4 | 15.6 | 13.2 |
| 5 | 6.3 | 5.1 |
| 6 | 8.2 | 6.3 |
| 7 | 13.4 | 10.2 |
| 8 | 15.2 | 11.3 |

As can be seen from Table 1, during the process of unordered target recognition by intelligent robots, the recognition time of the proposed method is significantly shorter than that of the method in Ref. [3], which fully demonstrates that the proposed method has a relatively high recognition efficiency. Compared with the method in Ref. [3], all the test indicators of the proposed method are superior, indicating that the proposed method can achieve a more satisfactory target image enhancement effect, laying a solid foundation for the subsequent unordered target recognition of intelligent robots.

## 4. Conclusion

Against the backdrop of the obvious deficiencies in traditional unordered target recognition methods for intelligent robots, in order to effectively address these issues, we propose a new unordered target recognition method for intelligent robots based on reinforcement learning. Specifically, first, the bilateral filtering technique is used to decompose the captured target images into low - illumination images and reflection images. Then, compression and enhancement processing are carried out on these two parts of the images respectively, and the processed images are merged thereafter. Finally, the merged images are input into the deep reinforcement learning model for training, so as to achieve the unordered target recognition of intelligent robots. The results of the simulation experiments show that this method performs excellently. It can not only achieve a satisfactory enhancement effect of target images, but also significantly improve the accuracy of the unordered target recognition results of intelligent robots and greatly shorten the recognition time, thus better meeting the various requirements of the system for the unordered target recognition of intelligent robots.